# Orientation Control with Variable Stiffness Dynamical Systems

Youssef Michel[1], Matteo Saveriano[2], Fares J. Abu-Dakka[3], and Dongheui Lee[4]

*Abstract*— Recently, several approaches have attempted to combine motion generation and control in one loop to equip robots with reactive behaviors, that cannot be achieved with traditional time-indexed tracking controllers. These approaches however mainly focused on positions, neglecting the orientation part which can be crucial to many tasks e.g. screwing. In this work, we propose a control algorithm that adapts the robot's rotational motion and impedance in a closed-loop manner. Given a first-order Dynamical System representing an orientation motion plan and a desired rotational stiffness profile, our approach enables the robot to follow the reference motion with an interactive behavior specified by the desired stiffness, while always being aware of the current orientation, represented as a Unit Quaternion (UQ). We rely on the Lie algebra to formulate our algorithm, since unlike positions, UQ feature constraints that should be respected in the devised controller. We validate our proposed approach in multiple robot experiments, showcasing the ability of our controller to follow complex orientation profiles, react safely to perturbations, and fulfill physical interaction tasks.

## I. INTRODUCTION

Robots nowadays are expected to transition from constrained, well-defined industrial settings, to more domestic environments such as museums and hospitals. Therefore, robots should be capable of operating in a wide range of environments, where they can safely co-exist in close proximity with people, and handle a wide array of tasks. This implies that robots should be endowed with *motion generators* to perform complex manipulation tasks, which potentially consist of position and orientation, and *controllers* that can track these motions, while being safe and reactive to possible perturbations.

Impedance control [1] is a well-established approach to shape the robot impedance behavior during physical interactions, thereby reducing the effect of position uncertainties and contact forces. More recently, Variable Impedance Control (VIC) [2] arised as a powerful tool where the impedance parameters over time or state are adapted to account for different task settings and environments. For example, the stiffness of the robot can vary depending on the external interaction forces that reflect friction or disturbance levels during autonomous operation [3] or during bilateral teleoperation [4]. With VIC, it is also possible to realize control behaviors that optimize some performance metrics such as the tracking error, interaction forces, and the metabolic cost, and where the control policy can be learnt with iterative learning control [5] or reinforcement learning [6]. Human-robot collaboration applications have also benefited from VIC, where the robot compliance can be shaped depending on the task stage [7], or the human intent during the incremental refinement of robotic skills [8].

Concerning the motion generation part, the Learning from Demonstration (LfD) paradigm is becoming increasingly popular thanks to its flexibility in learning motion plans based on human demonstrations without the need for re-programming, making it particularly appealing for non-experts. Several approaches have been proposed for encoding demonstrations, such as Gaussian Mixture Models (GMM), hidden Markov models [9], and dynamic movement primitives [10]. Among the several approaches, first-order Dynamical Systems (DS) have gained interest in modeling robot motions thanks to their stability properties, their time invariance (which makes them robust to temporal perturbations), and their flexibility to employ with different machine learning algorithms [11], [12]. DS have been also deployed to tasks such as catching flying objects [13], avoiding obstacles [14], hybrid motion-force control [15], and shared control [16].

The common convention in most of the existing literature is to treat motion generation and control as two separate problems [17], [18]. This gave rise to the traditional open-loop architecture, whereby the motion generator is completely isolated from the control loop, i.e., the motion planner is used to generate a time-indexed trajectory for the controller to follow. This means that the planner is agnostic to the current robot state and it is unable to react to unexpected dangerous situations (e.g., a robot is getting stuck behind an obstacle resulting in a clamping scenario [19]).

To solve this problem, several works have attempted to combine motion generation and control in one loop, eliminating the notion of "tracking" a trajectory. For instance, [20] proposes to encode motion and impedance behaviors in a potential field learnt from human demonstrations. This strategy is inspired by the popular potential field approach [21]. In [17], the authors reformulate a GMM into a closed-loop control policy with a state-varying spring and a damper, while [22] proposes to follow a path encoded in a velocity field, with a virtual fly-wheel used to ensure the passivity of the overall system. Early work in DS-based motion planning exploited the open-loop formulation, i.e., they generate the reference trajectory by open-loop integrating the DS. This was remedied in [18], where the authors proposed a passive controller that follows a velocity field in closed-loop. However, the controller in [18] lacks the ability to follow a particular demonstrated path in order to converge to the goal, which was solved in [23] by encoding a symmetric attraction behavior around a reference path. In our previous work [24], we proposed Variable Stiffness Dynamical Systems (VSDS), to further encode variable stiffness behaviors in a control policy that is able to follow a first-order DS in closed-loop, and with a symmetric attraction behavior similar to [23].

[1]Chair of Human-Centered Assistive Robotics, Technical University of Munich, Munich, Germany `youssef.abdelwadoud@tum.de`
[2]Department of Industrial Engineering, University of Trento, Trento, Italy `matteo.saveriano@unitn.it`
[3]Munich Institute of Robotics and Machine Intelligence, Technical University of Munich, Munich, Germany `fares.abu-dakka@tum.de`
[4]Autonomous Systems, TU Wien, Vienna, Austria and the Institute of Robotics and Mechatronics, German Aerospace Center (DLR), Munich, Germany `dongheui.lee@tuwien.ac.at`
This work has been partially supported by the European Union under the NextGenerationEU project iNest (ECS 00000043) and by euROBIN under grant agreement No 101070596.

The aforementioned works focus on developing closed-loop motion generation and control architectures targeted for position tasks, with no-regard to the robot orientation, which can be crucial in achieving tasks such as screwing or wiping on a curved surface. In this work, we develop a control algorithm that simultaneously shapes the robot orientation motion and rotational impedance, in a closed-loop manner that is always aware of the current robot orientation. Thereby, we extend the reactivity and safety properties of closed-loop motion generators and control to the orientation case. To achieve that, we extend our VSDS framework [24] to handle orientations, therefore combining the flexibility of DS motion generators with the interaction abilities of VIC for orientation tasks. Our framework relies on Unit Quaternions (UQ), a singularity-free representation of orientations. However, this poses a challenge since UQ geometric constraints (i.e., unitary norm) must be respected in the control formulation. This is a key difference with respect to position trajectories, which can be modeled in a decoupled manner. We overcome this problem by relying on the Lie group formulation and leveraging the mappings between the unit sphere and the tangent space to respect the underlying structure of UQ.

## II. PRELIMINARIES

### A. Original VSDS

In our previous work [24], we presented the VSDS control algorithm. The main idea of VSDS is to devise a control force that concurrently controls the robot's motion and interaction behavior. For the desired motion, we assume that it is represented as a first-order DS that can be learnt with any state-of-the-art approaches, e.g., [11], [12]. The DS is constructed as $\dot{x}_d = f_g(x)$ which maps the current linear position $x$ into a desired velocity $\dot{x}_d$. Typically, the way the robot is programmed to follow $\dot{x}_d$ is to generate a time-indexed trajectory $x_d(t)$ by successively integrating $\dot{x}_d$ starting from the initial position, which clearly constitutes an open-loop configuration. In VSDS, our goal is also to follow $f_g(x)$, however in a closed-loop configuration and with a symmetric attraction behavior around a path described by one of the integral curves of $f_g(x)$. The interaction behavior around the path is specified by a user-defined (variable) stiffness profile $K_p(x)$, which can be obtained via, e.g., regression-based approaches [3], [4] or through an external modality such as EMG [25]. The VSDS is constructed as a non-linear weighted combination of local springs, each constructed around a local attractor sampled from $f_g(x)$. Fig. 1 left shows example streamlines for an original DS learnt with SEDS [11]; an algorithm that learns asymptotically stable first-order DS represented as Gaussian Mixture Models. Fig. 1 right shows the streamlines of VSDS constructed from the same DS.

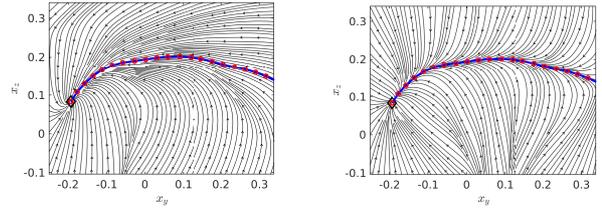

Fig. 1. Left: Streamlines of a 1st-order DS learnt with the Stable Estimator of Dynamical Systems (SEDS) [11]. Right: Streamlines of VSDS constructed from the same 1st-order DS. In both, the rhombus is the goal, the blue line is the reference path starting from a certain initial position, while the red dots are the computed via-points.

To extend VSDS to orientations, a suitable representation for the orientation DoF that does not suffer from representation singularities is UQ. Since UQ do not belong to Euclidean space, applying the standard arithmetic tools from Euclidean geometry would result in severe distortions and inaccuracies. This can be solved by exploiting proper mathematical tools developed for manifolds, as will be discussed next.

### B. Unit Quaternion Group

A quaternion is an element of $\mathbb{H}$, where $\mathbb{H}$ is the space of quaternions and it is isomorph to $\mathbb{R}^4$. A quaternion can be defined using a hyper-complex number as $\mathbf{q} = \nu + \mathbf{u} : \nu \in \mathbb{R}, \mathbf{u} = [u_x, u_y, u_z]^\top \in \mathbb{R}^3$. A UQ $\mathbf{q} \in \mathcal{S}^3$ is an element of $\mathcal{S}^3$, i.e., the unit sphere embedded in $\mathbb{R}^4$. A UQ is a quaternion with a unit norm and can be used to describe an orientation in 3D space, where $\mathbf{q}$ and $-\mathbf{q}$ represent the same orientation. Quaternion norm is obtained by $||\mathbf{q}|| = \sqrt{\nu^2 + u_x^2 + u_y^2 + u_z^2}$. The conjugation of a quaternion $\mathbf{q}$ is defined as $\bar{\mathbf{q}} = \nu + (-\mathbf{u})$, while the multiplications of $\mathbf{q}_1, \mathbf{q}_2 \in \mathcal{S}^3$ is defined as

$$\mathbf{q}_1 * \mathbf{q}_2 = (\nu_1 \nu_2 - \mathbf{u}_1^\top \mathbf{u}_2) + (\nu_1 \mathbf{u}_2 + \nu_2 \mathbf{u}_1 + \mathbf{u}_1 \times \mathbf{u}_2).$$

We project UQ back and forth between the unit sphere manifold $\mathcal{S}^3$ and the tangent space $\mathbb{R}^3$ by using exponential and logarithmic mapping operators. $\boldsymbol{\zeta} = \mathrm{Log}_{\mathbf{q}_2}(\mathbf{q}_1) : \mathcal{S}^3 \mapsto \mathcal{T}_{\mathbf{q}_2} \mathcal{S}^3$; $\mathcal{T}_{\mathbf{q}_2} \mathcal{S}^3 \equiv \mathbb{R}^3$, maps $\mathbf{q}_2 \in \mathcal{S}^3$ to $\boldsymbol{\zeta} \in \mathbb{R}^3$ wrt to $\mathbf{q}_1$. Consider $\mathbf{q} = \mathbf{q}_2 * \bar{\mathbf{q}}_1$,

$$\mathrm{Log}_{\mathbf{q}_1}(\mathbf{q}_2) = \mathrm{Log}(\mathbf{q}) = \begin{cases} \arccos(\nu) \frac{\mathbf{u}}{||\mathbf{u}||}, & ||\mathbf{u}|| \neq 0 \\ [0\ 0\ 0]^\top, & \text{otherwise} \end{cases} \quad (1)$$

This mapping can be also used to define a distance metric upon $\mathcal{S}^3$ [26]

$$\mathrm{d}(\mathbf{q}_1, \mathbf{q}_2) = \begin{cases} 2\pi, & \mathbf{q}_1 * \bar{\mathbf{q}}_2 = [-1 + [0\ 0\ 0]^\top] \\ 2||\mathrm{Log}_{\mathbf{q}_1}(\mathbf{q}_2)||, & \text{otherwise} \end{cases}.$$

(2)

Inversely, $\mathbf{q} = \mathrm{Exp}_{\mathbf{q}_1}(\boldsymbol{\zeta}) : \mathbb{R}^3 \mapsto \mathcal{S}^3$ maps $\boldsymbol{\zeta} \in \mathbb{R}^3$ to $\mathbf{q} \in \mathcal{S}^3$ so that it lies on the geodesic starting point from $\mathbf{q}_1$ in the

direction of $\boldsymbol{\zeta}$

$$\text{Exp}_{\mathbf{q}_1}(\boldsymbol{\zeta}) = \begin{cases} \left[\cos(||\boldsymbol{\zeta}||) + \sin(||\boldsymbol{\zeta}||)\frac{\boldsymbol{\zeta}}{||\boldsymbol{\zeta}||}\right] * \mathbf{q}_1, & ||\boldsymbol{\zeta}|| \neq 0 \\ [1 + [0\ 0\ 0]^\top] * \mathbf{q}_1, & \text{otherwise} \end{cases} \quad (3)$$

It is worth mentioning that UQs are often computed numerically from rotation matrices, for instance when robot's end-effector poses are collected via kinesthetic teaching. In these cases, it may happen that the algorithm returns a quaternion at step $t$ and an antipodal quaternion at $t+1$. To ensure that the demonstration of an orientation profile is discontinuity-free, we can check that the dot product between each adjacent UQ is greater than zero. Otherwise, we flip $\mathbf{q}_{t+1}$ such as $\mathbf{q}_{t+1} = -\mathbf{q}_{t+1}$.

*Remark 1:* As discussed in [27], the domain of $\text{Log}(\cdot)$ extends to all $\mathcal{S}^3$ except $-1 + [0\ 0\ 0]^\top$, while the domain of $\text{Exp}(\cdot)$ is constrained by $||\boldsymbol{\zeta}|| < \pi$. Restricting the domain to $||\boldsymbol{\zeta}|| < \pi$ makes (1) and (3) bijective.

## III. Proposed Approach

In this section, we formulate the VSDS algorithm for the orientation case. Similar to the classical VSDS [24], we assume that there exists a known nominal motion plan represented by a first-order dynamical system $\boldsymbol{\omega}_d = \boldsymbol{f}_g(\boldsymbol{q})$ asymptotically stable around a global equilibrium $\boldsymbol{q}^*$, with $\boldsymbol{f}_g(\cdot) : \mathcal{S}^3 \mapsto \mathbb{R}^3$ a continuously differentiable function that maps a UQ $\boldsymbol{q} \in \mathcal{S}^3$ into a desired angular velocity $\boldsymbol{\omega}_d \in \mathbb{R}^3$. As shown in [28], [29], the function $\boldsymbol{f}_g(\cdot)$ can be learned from demonstration by projecting quaternion trajectories in the tangent space using the logarithmic map (1). We also have a desired, possibly state-varying stiffness profile $\boldsymbol{K}_o(\mathbf{q}) \in \mathbb{R}^{3 \times 3}$ which describes the desired interaction behavior for the orientation degrees of freedom.

In the following, we outline the key elements of the VSDS algorithm for orientations represented as UQs. As discussed earlier, applying operations from Euclidean geometry directly on UQ leads to inaccuracies. Therefore, we make consistent use of projections back and forth between the manifold and the tangent space, since elements of the tangent space have a Euclidean geometry, and accordingly, we can exploit the well-known linear algebra and arithmetic tools to manipulate its elements.

### A. Via Points Extraction

The first step in our VSDS algorithm is to obtain a sequence of $N$ via-points generated from one of the integral curves of $\boldsymbol{f}_g(\boldsymbol{q})$ starting at an initial orientation $\boldsymbol{q}_0$. These via-points act as attractors for the local springs, thereby shaping the motion, and are chosen to be equidistant to ensure a smooth velocity profile. The via-points extraction process is detailed in Algorithm 1, where first the motion plan DS is simulated with a sampling time $\Delta t$ until reaching $\boldsymbol{q}^*$, generating a temporary sequence of via-points, which are then projected in the tangent space at the goal quaternion $\boldsymbol{q}^*$ (Lines 7-13). The temporary via-points are then re-sampled to obtain $N$ equidistant via-points $\boldsymbol{\zeta}_{l,0}, \boldsymbol{\zeta}_{l,i}, \dots \boldsymbol{\zeta}_{l,N}$ in the tangent space and $N$ equidistant via-points $\boldsymbol{q}_{l,0}, \boldsymbol{q}_{l,i}, \dots \boldsymbol{q}_{l,N}$ in the unit quaternion space (Lines 15-22).

---

**Algorithm 1:** SampleViaPoints

**input :** Motion plan DS $\boldsymbol{f}_g$, initial $\boldsymbol{q}_0$ and goal orientation $\boldsymbol{q}^*$, Number of via-points $N$, sampling time $\Delta t$

1 $n = 0$;
2 $k = 1$;
3 $d_{sum} = 0$;
4 $\boldsymbol{\zeta}_{l,n} = \text{Log}_{\boldsymbol{q}^*}(\boldsymbol{q}_0)$;
5 $\boldsymbol{\zeta}_{tmp,k} = \text{Log}_{\boldsymbol{q}^*}(\boldsymbol{q}_0)$;
6 $\boldsymbol{q}_{tmp,k} = \boldsymbol{q}_0$;
7 **while** $\text{d}(\boldsymbol{q}_k, \boldsymbol{q}^*) > \epsilon$ **do**
8 $\quad$ $k = k + 1$;
9 $\quad$ $\boldsymbol{\zeta}_{tmp,k} = \text{Log}_{\boldsymbol{q}^*}(\boldsymbol{q}_{tmp,k})$ ;
10 $\quad$ $\boldsymbol{\omega}_{tmp,k} = \boldsymbol{f}_g(\boldsymbol{q}_{tmp,k})$ ;
11 $\quad$ $\boldsymbol{q}_{tmp,k+1} = \text{Exp}_{\boldsymbol{q}^*}(\frac{\Delta t}{2}\boldsymbol{\omega}_{d,k}) * \boldsymbol{q}_{tmp,k}$ ;
12 $\quad$ $d_{sum} = d_{sum} + \text{d}(\boldsymbol{q}_{tmp,k}, \boldsymbol{q}_{tmp,k-1})$ ;
13 **end**
14 $d_l = d_{sum}/N$;
15 **for** $i \leftarrow 1$ **to** $k$ **do**
16 $\quad$ **if** $\text{d}(\boldsymbol{q}_{tmp,i}, \boldsymbol{q}_{k,n}) \geq d_l$ **then**
17 $\quad\quad$ $n = n + 1$;
18 $\quad\quad$ $\boldsymbol{q}_{l,n} = \boldsymbol{q}_{tmp,i}$;
19 $\quad\quad$ $\boldsymbol{\zeta}_{l,n} = \boldsymbol{\zeta}_{tmp,i}$;
20 $\quad\quad$ $\boldsymbol{\omega}_{l,n} = \boldsymbol{\omega}_{tmp,i}$ ;
21 $\quad$ **end**
22 **end**
23 $\boldsymbol{\zeta}_{l,N} = \text{Log}_{\boldsymbol{q}^*}(\boldsymbol{q}^*) = [0\ 0\ 0]^\top$, $\boldsymbol{q}_{l,N} = \boldsymbol{q}^*$;

---

### B. Local Springs

The goal now is to construct $N$ local spring actions around the generated via-points which serve as local attractors for each spring. The spring actions are realized by a DS represented in the tangent space, such that

$$\boldsymbol{f}_{l,i}(\boldsymbol{\zeta}) = \boldsymbol{A}_{l,i}\text{Log}_{\boldsymbol{q}_{l,i}}(\boldsymbol{q}) \quad (4)$$

where $\text{Log}_{\boldsymbol{q}_{l,i}}(\boldsymbol{q})$ is the projection of the current robot orientation $\boldsymbol{q}$ in the tangent space at the via-point $\boldsymbol{q}_{l,i}$ (and it plays the role of an error term). Please note that we write $\boldsymbol{f}_{l,i}(\boldsymbol{\zeta})$ to highlight the fact that $\boldsymbol{f}_{l,i}$ maps elements that belong to the tangent space. The stiffness of each spring is $\boldsymbol{A}_{l,i}$, which is constructed based on the desired stiffness profile $\boldsymbol{K}_o$, and designed as

$$\boldsymbol{A}_{l,i} = \boldsymbol{Q}_{l,i}\boldsymbol{K}_{o,i}\boldsymbol{Q}_{l,i}^\top, \quad (5)$$

where $\boldsymbol{K}_{o,i} = \boldsymbol{K}(\boldsymbol{q}_{l,i})$ is the value of the stiffness matrix evaluated at the current via-point $\boldsymbol{q}_{l,i}$, while $\boldsymbol{Q}_{l,i}$ is a matrix constructed to have its first eigenvector normalized and pointing in the direction of motion, i.e., $\frac{\boldsymbol{\zeta}_{l,i}}{||\boldsymbol{\zeta}_{l,i}||}$ whereas the remaining vectors are derived to be perpendicular to it, via a Gram-Schmit orthogonalization procedure. Intuitively, the projection in (5) interprets the first eigenvalue of $\boldsymbol{K}_o$ to be the stiffness along the direction of motion, while the remaining eigenvalues as the stiffness perpendicular to the current motion direction.

Once the springs are constructed, we combine them via a non-linear weighted sum. The weights act as a state-dependent activation magnifying or reducing the effect of each spring depending on the region of the state space, while achieving a smooth transition. As in the original VSDS, we generate this smooth transition using a Gaussian kernel. A Gaussian kernel on the unit sphere can be defined as [30]

$$w_i(\boldsymbol{\zeta}) = \exp\left(-\frac{\text{Log}_{\boldsymbol{q}_{cen,i}}(\boldsymbol{q})^\top \text{Log}_{\boldsymbol{q}_{cen,i}}(\boldsymbol{q})}{2(\sigma^i)^2}\right) \quad (6)$$

where $\boldsymbol{q}_{cen,i}$ denotes the center of the $i_{th}$ DS, i.e., the mean of $\boldsymbol{q}_{l,i}$ and $\boldsymbol{q}_{l,i-1}$, and it can be efficiently computed using the EM-based approach outlined in [30]. $\sigma^i \in \mathbb{R}^+$ is proportional to the distance between two tangent spaces via points, computed using (2). The actual spring weights used are then the normalized kerenels $\tilde{w}_i(\boldsymbol{\zeta}) = \frac{w_i(\boldsymbol{\zeta})}{\sum_{j=1}^N w_j(\boldsymbol{\zeta})}$.

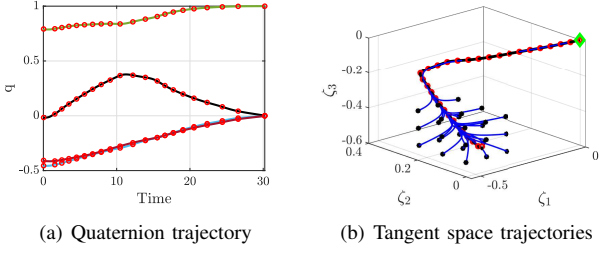

(a) Quaternion trajectory   (b) Tangent space trajectories

Fig. 2. Left: reference path for VSDS as a UQ where the 4 lines correspond to the $v$, $u_x$, $u_y$ and $u_z$. Right: shows the state trajectories of a second-order DS driven by VSDS starting from multiple initial orientations (black dots). The black line is the reference path, while the green rhombus is the goal. In both plots, the red dots are the computed via-points.

### C. VSDS Control policy

Finally, our VSDS control policy is devised as the weighted combination of all local springs, such that

$$\boldsymbol{\tau}_{vs}(\boldsymbol{q}) = \sum_{i=1}^N \tilde{w}_i(\boldsymbol{\zeta}) \boldsymbol{f}_{l,i}(\boldsymbol{\zeta}) \quad (7)$$

which represents a non-linear torque field that maps the current orientation $\boldsymbol{q}$ (computed through $\boldsymbol{\zeta}$) into a torque $\boldsymbol{\tau}_{vs} \in \mathbb{R}^3$ that simultaneously shapes the robot motion and rotational stiffness, in closed loop. Fig. 2(a) shows the computed via-points (red-dots) along a quaternion trajectory simulated from a first-order DS learnt with [29], while Fig. 2(b) visualizes the same DS in the tangent space at the goal $\boldsymbol{q}^*$ i.e $\mathcal{T}_{\boldsymbol{q}^*}\mathcal{S}^3$, and where also depicted the evolution of a rotational second order dynamical system driven by VSDS starting from multiple initial positions. Please note that all trajectories converge to the reference path along which the via-points are computed, thanks to the symmetric attraction behavior encoded within VSDS. The entire procedure for the VSDS approach including the initialization and the real-time control is outlined in Algorithm 2.

The Cartesian control torque for the orientation degrees of freedom can then be set as

$$\boldsymbol{\tau} = \boldsymbol{\tau}_{vs} - \boldsymbol{D}_o \boldsymbol{\omega} \quad (8)$$

**Algorithm 2:** Orientation Control with VSDS

**input :** $\boldsymbol{f}_g$, $\boldsymbol{q}^*$, $\boldsymbol{q}_0$, $\Delta t$, $N$, Current robot orientation $\boldsymbol{q}$ and Desired stiffness profile $\boldsymbol{K}_o(\boldsymbol{q})$

VSDS Initialization:
1. Algorithm 1 ;
2. **for** $i \leftarrow 1$ **to** $N$ **do**
3. $\quad \boldsymbol{K}_{o,i} = \boldsymbol{K}_o(\boldsymbol{q}_{l,i})$ ;
4. $\quad \boldsymbol{A}_{l,i} = \boldsymbol{Q}_i \boldsymbol{K}_{o,i} \boldsymbol{Q}_i^T$;
5. $\quad \boldsymbol{q}_{cen,i}$ mean of $\boldsymbol{q}_{l,i}$ and $\boldsymbol{q}_{l,i-1}$ [30];
6. $\quad l_i = \text{d}(\boldsymbol{q}_{l,i}, \boldsymbol{q}_{l,i-1})$;
7. $\quad \sigma_i = \delta l_i$;
8. **end**

Control Loop:
9. **while** $\text{d}(\boldsymbol{q}, \boldsymbol{q}^*) > \epsilon$ **do**
10. $\quad$ **for** $i \leftarrow 1$ **to** $N$ **do**
11. $\quad\quad w_i(\boldsymbol{\zeta}) = \exp\left(-\frac{\text{Log}_{\boldsymbol{q}_{cen,i}}(\boldsymbol{q})^\top \text{Log}_{\boldsymbol{q}_{cen,i}}(\boldsymbol{q})}{2(\sigma^i)^2}\right)$;
12. $\quad$ **end**
13. $\quad$ **for** $i \leftarrow 1$ **to** $N$ **do**
14. $\quad\quad \boldsymbol{f}_{l,i}(\boldsymbol{\zeta}) = \boldsymbol{A}_{l,i} \text{Log}_{\boldsymbol{q}_{l,i}}(\boldsymbol{q})$;
15. $\quad\quad \tilde{w}_i(\boldsymbol{\zeta}) = \frac{w_i}{\sum_{j=1}^N \omega_j(\boldsymbol{\zeta})}$;
16. $\quad$ **end**
17. $\quad \boldsymbol{\tau}_{vs}(\boldsymbol{q}) = \sum_{i=1}^N \tilde{w}_i(\boldsymbol{\zeta}) \boldsymbol{f}_{l,i}(\boldsymbol{\zeta})$ ;
18. **end**

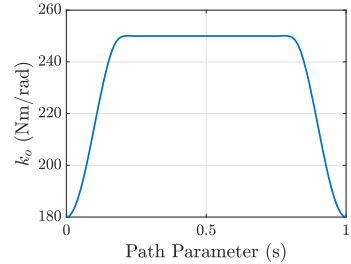

Fig. 3. Variable Stiffness Profile

where $\boldsymbol{\omega}$ is the actual angular velocity of the robot, while $\boldsymbol{D}_o$ is a damping matrix. To command the robot joints, the actuation torques sent to the motors are computed as

$$\mathbf{u} = \boldsymbol{J}^\top \begin{bmatrix} \boldsymbol{F} \\ \boldsymbol{\tau} \end{bmatrix}, \quad (9)$$

with $\boldsymbol{J}$ as the end-effector jacobian, and where the robot is assumed to be pre-gravity compensated. As for $\boldsymbol{F} \in \mathbb{R}^3$, it is a generic control force for the translational degrees of freedom, which can be commanded also via VSDS as in [24], or with a standard impedance controller.

## IV. EXPERIMENTAL VALIDATION

In this section, we aim to validate our approach in a series of experiments on a 7 DoF KUKA LWR robot. We implemented our control algorithm on a standard core i7 PC with 16 GB RAM, in C++. The robot was controlled via the Fast Research Interface in the joint impedance control mode, where the feed-forward torque command is used for our VSDS control law. The frequency of the control loop runs at 500 Hz. In the following, we present a series of experiments that highlight the ability of our VSDS to follow complex orientation profiles, safety during disturbances, and finally, an interaction task with varying stiffness levels[1]. For all experiments, we used $N = 30$ as the number of local spring actions. In order to generate a first-order DS, we used the Riemannian manifold stable DS (SDS-RM) [29] to learn the orientation motion plan DS $f_g$ from demonstrations. The idea of SDS-RM is to encode demonstrations that belong to a Riemannian manifold into a first-order stable DS. This is achieved by learning a diffeomorphism; a bijective smooth mapping with a continuous inverse; between tangent spaces to map a globally asymptotically stable base DS into a DS that can model complex motions, while preserving the stability properties of the base DS. Since UQ belong to Riemannian manifolds, SDS-RM can be adequately deployed to model demonstrated orientation trajectories.

*A. Free Motion Evaluation*

In this subsection, we test the ability of our control law to follow a desired orientation path. To that end, we use the Riemannian LASA Data Set [29], [28] to obtain orientation demonstration data from three shapes: JShape, Worm, and a Trapezoid. We then use SDS-RM [29] to learn a first-order DS, which is subsequently used to generate the via points for VSDS. For the stiffness, we used a variable stiffness profile parameterized as a function of the path parameter $s \in [0, 1]$ such that $s = 0$ at $q = q_0$ and $s = 1$ at $q = q^*$. The stiffness profile is designed to start with stiffness of 180 Nm/rad, rise smoothly to 250 Nm/rad and fall again back to 180 Nm/rad toward the end of the motion, as shown in Fig. 3, such that $\boldsymbol{K}_o = k_o(s)\boldsymbol{I}_3$. The translational degrees of freedom of the robot were commanded with a Cartesian impedance controller to hold the initial position.

The results of this experiment are shown in Figure 4, where depicted the actual robot (solid black) and desired (dotted red) orientation motions for the three shapes. The 3D plots in Figures 4(d), 4(e) and 4(f) highlight the evolution of the orientation trajectory projected in the tangent space $\mathcal{T}_{\mathbf{q}^*}\mathcal{S}^3$, while Figures 4(a), 4(b) and 4(c) show the evolution of the scalar and vector parts of the quaternion trajectories. The desired motion is generated through the open-loop integration of the learnt first order DS $\boldsymbol{f}_g(\boldsymbol{q})$, which represents the nominal motion plan. As highlighted in the figure, the actual robot orientation follows well the desired orientation profile. Please note, to ease visualization, the desired quaternion trajectory generated by the first order DS is scaled in time, in order to make it in the same time scale as the actual one.

*B. Safety*

We further validate the safety and robustness of our control algorithm by testing its ability to cope with various disturbances. The robot has to execute the J-Shape motion in a similar setting to the previous subsection, however, with a human applying perturbations by physically interacting with the robot. In the first case, the human attempts to stop the robot's motion, simulating a potential collision or an obstacle that might block the robot. As shown in the videos, the robot reacts in a compliant manner and resumes the task smoothly once the human releases the robot. This is also reflected in the VSDS control torques norm shown in Fig. 5(a), which remains almost constant during the period the human is holding the robot, highlighted by the two dotted vertical lines. This smooth behavior can be realized thanks to the closed-configuration, where the controller and motion generator are encoded as a single entity, that is always aware of the current robot state. On the other hand, a time-driven controller will result in a control torque that progressively increases with time, resulting in an abrupt and potentially dangerous behavior once the obstacle is removed, as shown for example in [18] and [24].

In the second case, the human repeatedly pushes the robot attempting to deviate it from its desired path. The robot again reacts to the applied disturbances compliantly, resuming the planned task by attracting back to the desired orientation path, as shown in the robot motions visualized in tangent space (Fig. 5(b)). This can be fulfilled thanks to the symmetric attraction behavior of VSDS, which is also consistent with the simulation results of Fig. 2(b).

*C. Interaction Task*

In this experiment, we validate the ability of our approach to execute physical interaction tasks. In particular, we chose a cutting task where a scalpel was mounted on the robot end-effector, and where the goal is to perform a curve-shaped cutting motion on air-drying modeling clay that naturally hardens over time. To achieve this objective, the orientation of the robot must continuously change in order to keep the cutting edge of the scalpel always aligned with the current motion direction. To learn the task, a human provides one demonstration to the robot via kinaesthetic teaching with the robot in gravity compensation mode where the robot pose data is collected. The recorded orientations are then used to learn a first-order DS with SDS-RM, which is used to generate the via-points for VSDS. During execution, we perform the same cutting motion in two different settings: In the first case, the robot has to cut a freshly opened piece of soft clay, and therefore we use a relatively low constant stiffness of $\boldsymbol{K}_o = 130\boldsymbol{I}_3$ Nm/rad to perform the cut. In the second case, we use a stiffness of $\boldsymbol{K}_o = 200\boldsymbol{I}_3$ Nm/rad to cut an older and comparatively harder material. For the translational motion, we simply play-back the recorded trajectory and track it with a Cartesian impedance controller. As shown in the attached video and in Fig. 6[2], the robot adapts its orientation based on the desired profile throughout the motion in order to prevent any lateral knife motions, and completes the task successfully.

---

[1] A video of the conducted experiments is provided in the supplementary material

[2] We show only the orientation trajectories of the hard material, since the trajectories of the soft are similar.

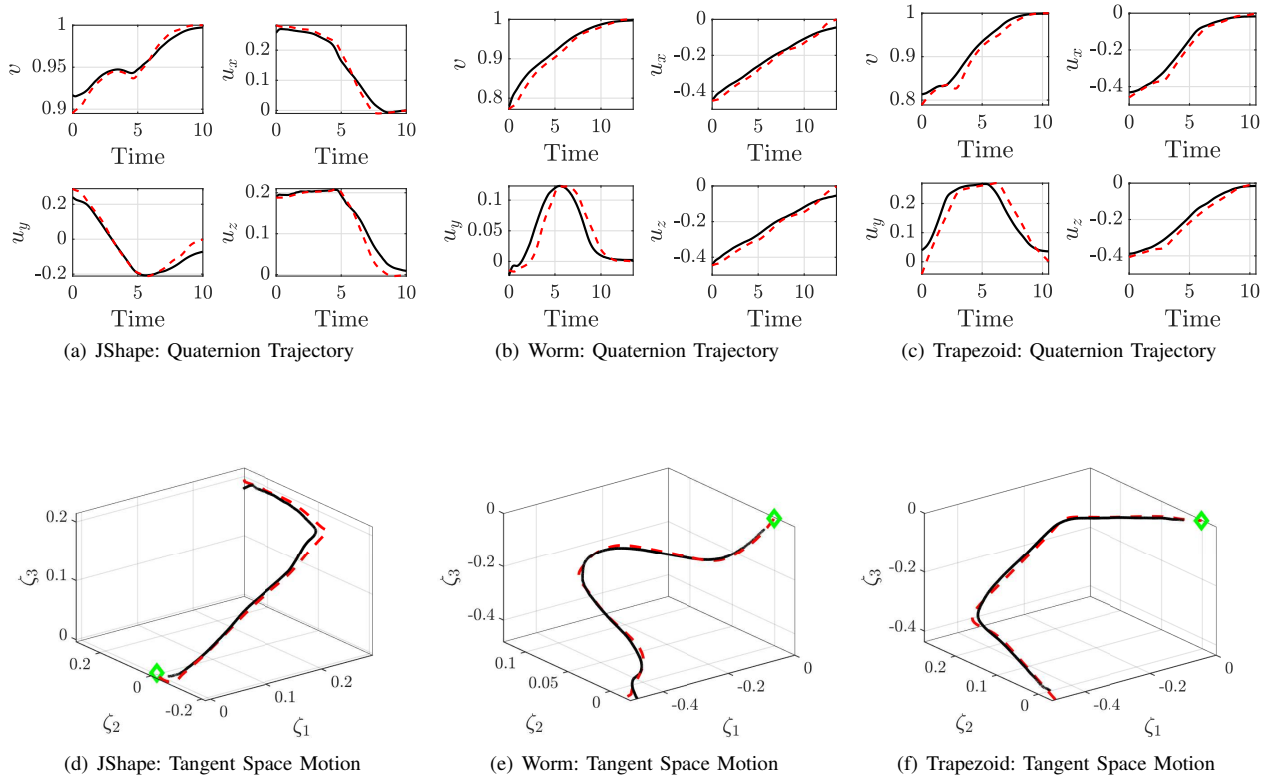

Fig. 4. Experimental results for free motion execution, where the black line depicts the actual robot motion, while the dotted red is the desired one.

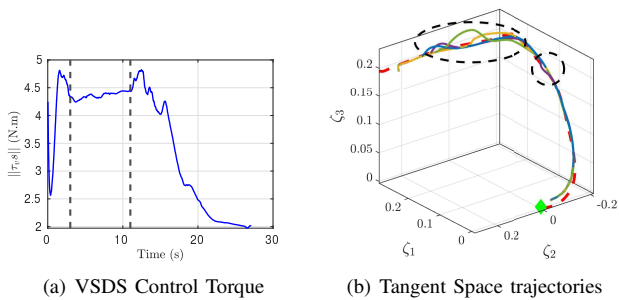

Fig. 5. Left: norm of the VSDS control torque during an execution where the human attempts to block the robot motion, where the black lines depict the duration of the interaction. Right: quaternion trajectories from several experiments where the human applies random disturbances to the robot. The dotted red is the reference path in the tangent space, while the dotted ellipsoids highlight the instances where the human applied the disturbances.

## V. CONCLUSIONS

In this work, we extended the VSDS approach for orientation tasks, to enable a robot to follow a desired orientation motion plan described by a first-order DS, with a user-defined rotational stiffness profile, in a closed-loop configuration capable of guaranteeing reactive robot behaviors. Our algorithm relies on UQ to represent orientations and projections between manifold and tangent space, to realize a control law formulated as a non-weighted sum of local springs centered around local attractors in the tangent space. Thanks to the chosen formulation, the controller also realizes a spring-like attraction behavior around a reference path, where the robot is pulled back to the path when perturbed. We validated during real-world experiments the ability of the proposed approach to follow complex orientation profiles and ensure safe and compliant reactions to perturbations. Finally, we conducted a cutting task where the robot orientation should continuously adapt to cut a curve shape, with two different stiffness levels.

In the future, along the same lines of [31], we will focus on studying more in-depth the asymptotic stability and passivity properties of the proposed controller. We will also endow our VSDS with trajectory-tracking capabilities in order to follow the desired orientation velocity profile, in addition to the orientation path.

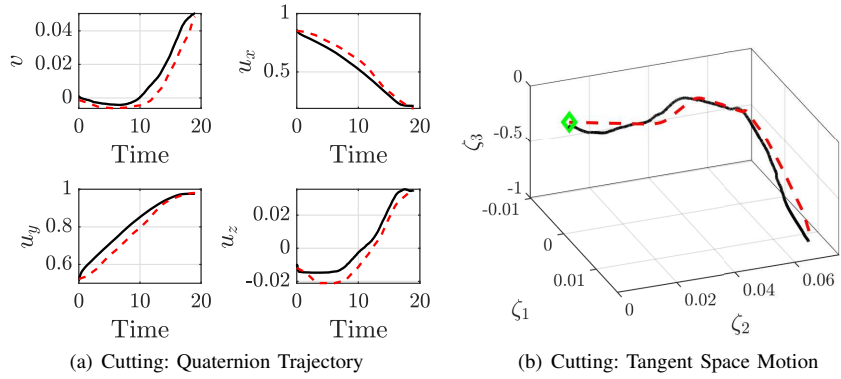
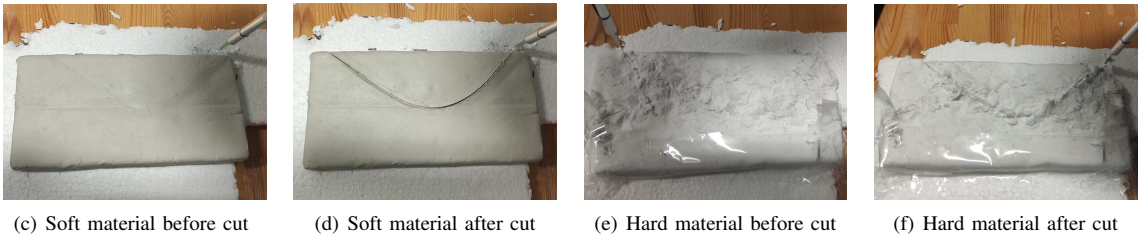

Fig. 6. Experimental results for Cutting